\pdfoutput=1

\documentclass[11pt]{article}

\usepackage{EMNLP2022}

\usepackage{times}
\usepackage{latexsym}

\usepackage[T1]{fontenc}

\usepackage[utf8]{inputenc}

\usepackage{microtype}

\usepackage{soul}
\usepackage{url}
\usepackage{graphicx}
\usepackage{amsmath}
\usepackage{amsthm}
\usepackage{booktabs}
\usepackage{algorithm}
\usepackage{algorithmic}

\usepackage{colortbl,booktabs}
\usepackage{threeparttable}
\usepackage{tabularx}
\usepackage{multirow}
\usepackage{multicol}
\usepackage{mathtools}
\usepackage{amsmath,amsfonts} 
\usepackage{subfigure}
\usepackage{bm}
\usepackage{xcolor}
\usepackage{color}
\usepackage{newfloat}
\usepackage{listings}

\title{Label-aware Multi-level Contrastive Learning for Cross-lingual \\ Spoken Language Understanding}

\author{
    Shining Liang$^{1,2,3}$\footnotemark[1], Linjun Shou$^{3}$, Jian Pei$^{4}$, Ming Gong$^{3}$, \\ \textbf{Wanli Zuo$^{1,2}$, Xianglin Zuo$^{1,2}$\footnotemark[2], Daxin Jiang$^3$\footnotemark[2]}\\
    $^1${College of Computer Science and Technology, Jilin University} \\
	$^2${Key laboratory of Symbolic Computation and Knowledge Engineering, MOE}\\
	$^3${STCA, Microsoft}\\
	$^4${Department of Electrical \& Computer Engineering, Duke University}\\
	{\small \{liangsn17,zuoxl17\}@mails.jlu.edu.cn;
	\{lisho,migon,djiang\}@microsoft.com; j.pei@duke.edu; zuowl@jlu.edu.cn}}

\begin{document}
\maketitle

\footnotetext[1]{Work is done during internship at Microsoft STCA.} 
\footnotetext[2]{Corresponding authors.}

\begin{abstract}
Despite the great success of spoken language understanding (SLU) in high-resource languages, it remains challenging in low-resource languages mainly due to the lack of labeled training data. 
The recent multilingual code-switching approach achieves better alignments of model representations across languages by constructing a mixed-language context in zero-shot cross-lingual SLU. However, current code-switching methods are limited to implicit alignment and disregard the inherent semantic structure in SLU, i.e., the hierarchical inclusion of utterances, slots, and words. 
In this paper, we propose to model the \emph{utterance-slot-word} structure by a multi-level contrastive learning framework at the utterance, slot, and word levels to facilitate explicit alignment. Novel code-switching schemes are introduced to generate hard negative examples for our contrastive learning framework. Furthermore, we develop a label-aware joint model leveraging label semantics to enhance the implicit alignment and feed to contrastive learning. Our experimental results show that our proposed methods significantly improve the performance compared with the strong baselines on two zero-shot cross-lingual SLU benchmark datasets.
\end{abstract}

\section{Introduction}
Spoken language understanding (SLU) is a critical component of goal-oriented dialogue systems, which consists of two subtasks: intent detection and slot filling~\cite{wang2005spoken}. 
Recently, massive efforts based on the joint training paradigm~\cite{chen2019bert,qin2021co} have shown superior performance in English. However, the majority of them require large amounts of labeled training data, which limits the scalability to low-resource languages with little or no training data. Zero-shot cross-lingual approaches have arisen to tackle this problem that transfer the language-agnostic knowledge from high-resource (source) languages to low-resource (target) languages. 

\begin{table}[t] \small
    \centering
        \begin{tabular}{l|cccc}
        \toprule
        \textbf{Method} & \textbf{en} & \textbf{es} & \textbf{zh} & \textbf{tr} \\
        \midrule
        zero-shot       & 88.24 & 52.18 & 30.01 & 3.08 \\
        code-switching  & 88.69 & 54.42 & 45.24 & 7.41 \\
        \bottomrule
        \end{tabular}
    \caption{mBERT based zero-shot and code-switching (CoSDA-ML) results on four languages of MultiATIS++ (semantic EM accuracy).}
    \label{tab:pre}
\end{table}

For the data-based transfer methods, machine translation is first applied to translate the source utterances into the targets~\cite{upadhyay2018almost,schuster2019cross,xu2020end}. However, machine translation may be unreliable or unavailable for some extremely low-resource languages~\cite{upadhyay2018almost}. Therefore, multilingual code-switching~\cite{liu2020attention,DBLP:conf/ijcai/QinN0C20} is developed to reduce the dependency on machine translation, which simply uses bilingual dictionaries to randomly select some words in the utterance to be replaced by the translation words in other languages. Code-switching has achieved promising results as the word representations in the mixed-language context are aligned in a universal vector space, which is essential for cross-lingual transfer~\cite{cao2020multilingual,chi2021infoxlm}.

\begin{figure*}[t]
    \centering
        \subfigure[Utterance Level]{
            \label{fig:ucl}
            \includegraphics[width=0.3\textwidth]{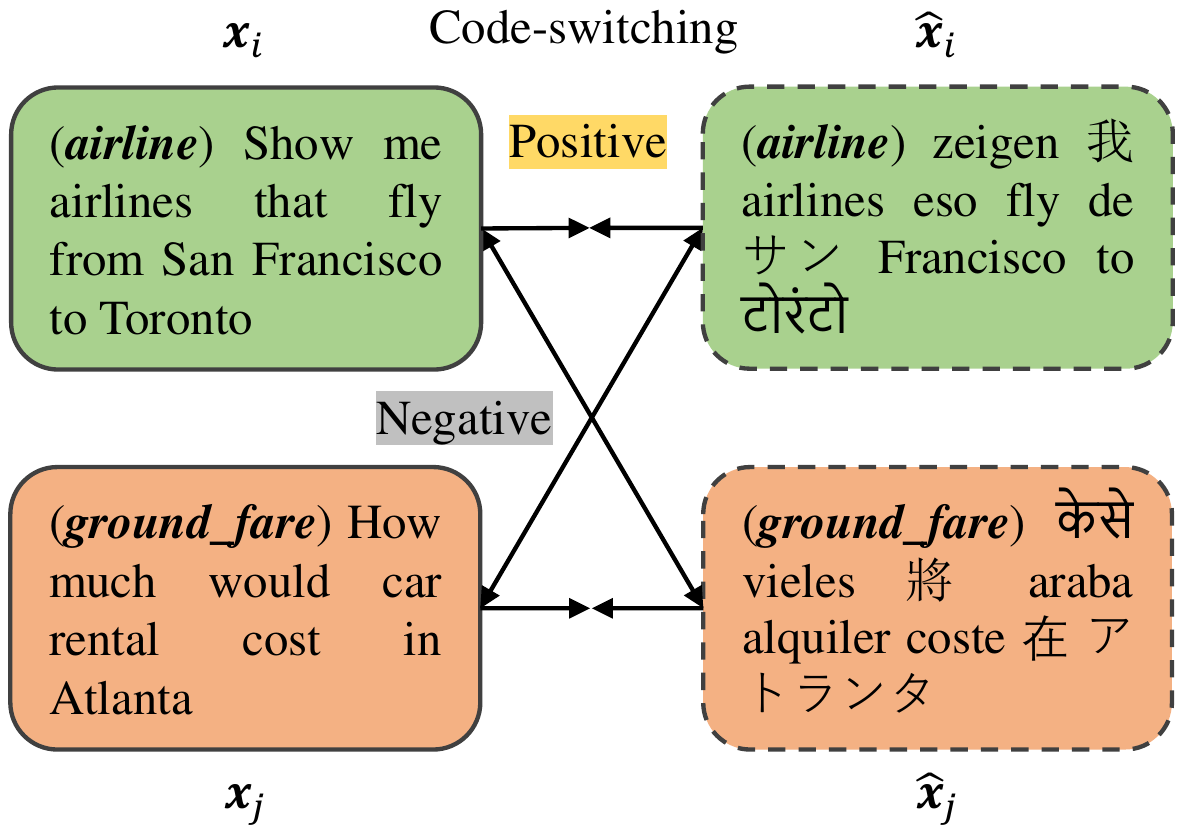}}
        \subfigure[Slot Level]{
            \label{fig:scl}
            \includegraphics[width=0.4\textwidth]{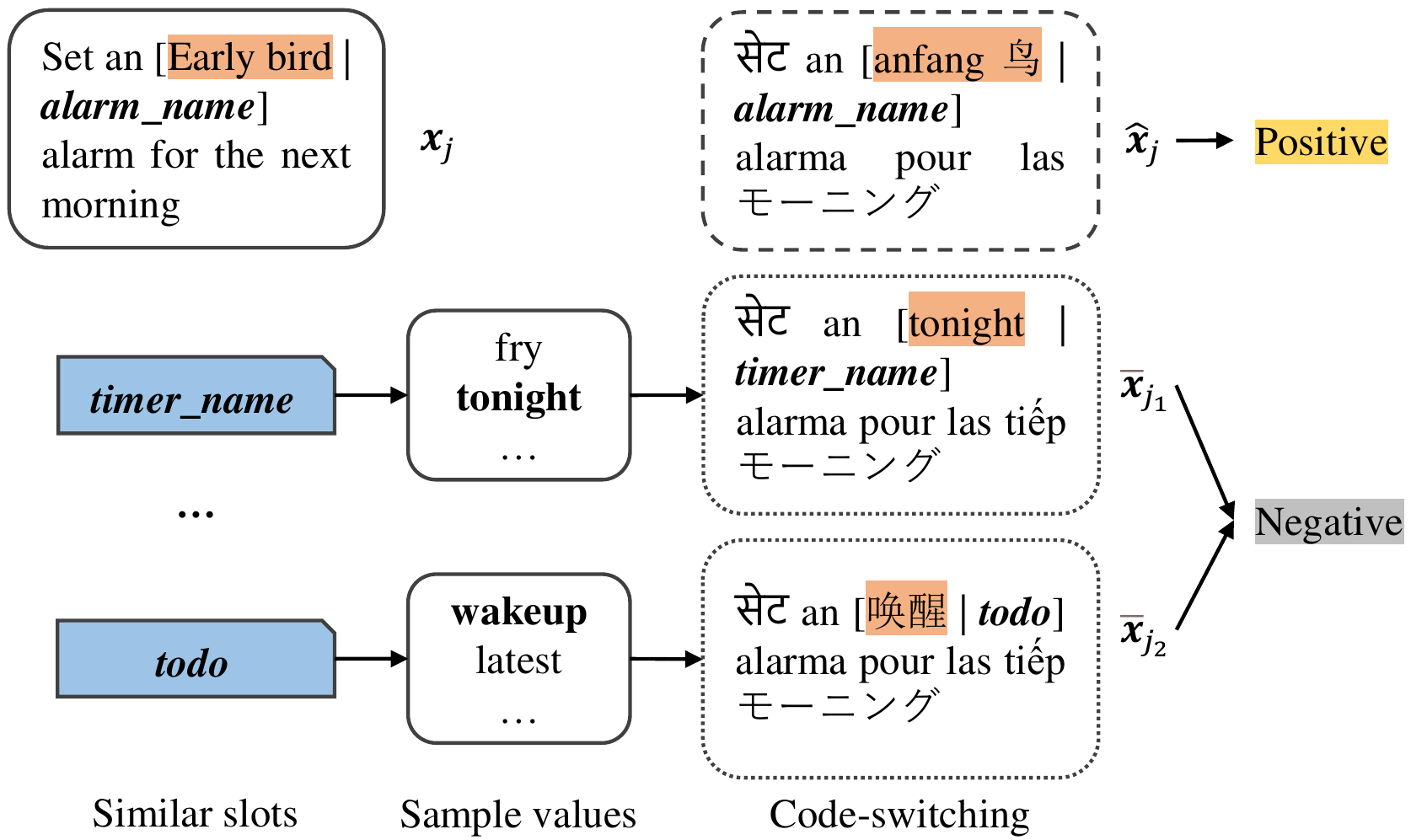}}
        \subfigure[Word Level]{
            \label{fig:wcl}
            \includegraphics[width=0.27\textwidth]{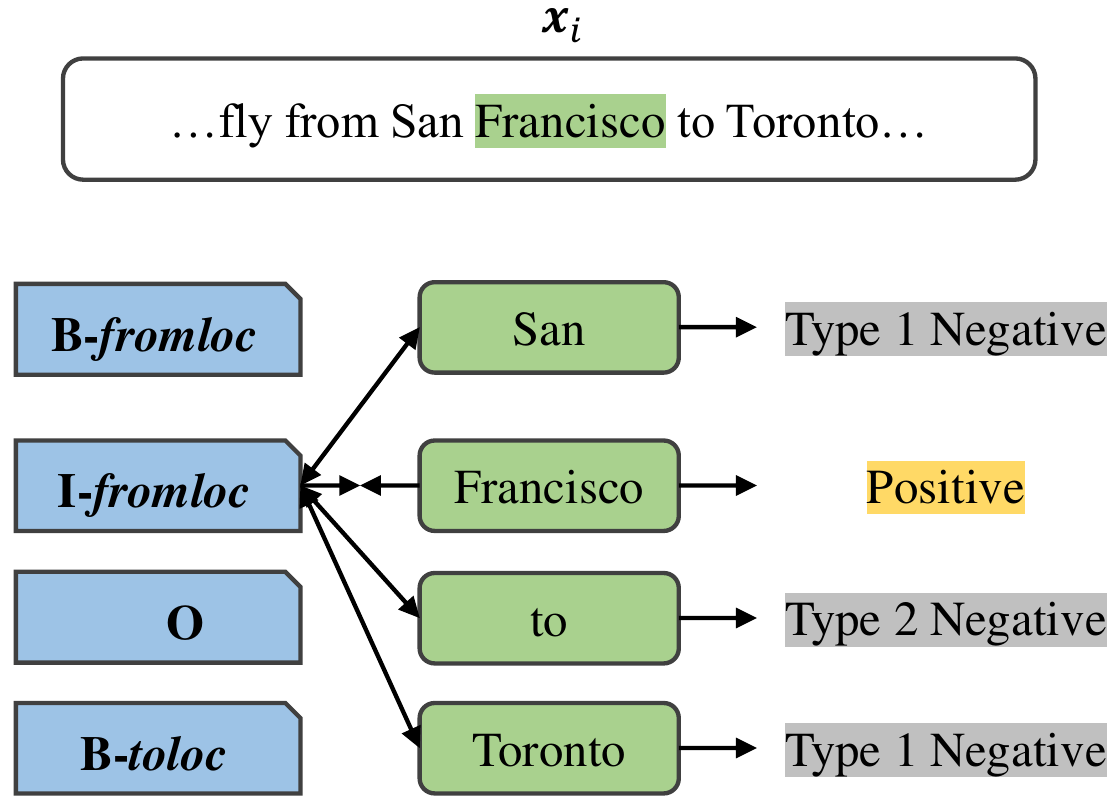}} \\
    \caption{Multi-level contrastive learning examples.}
    \label{fig:multi-level}
\end{figure*}

Despite the substantial improvements of CoSDA-ML~\cite{DBLP:conf/ijcai/QinN0C20} in Table~\ref{tab:pre}, there still exists a challenging performance gap between English and the target languages. We believe only the implicit alignment of code-switching is insufficient for refining model representations.
To address this issue, we advocate a fundamental methodology -- exploiting structures of utterances. In general, given a user utterance, there is a natural hierarchical structure, \emph{utterance-slot-word}, which describes the complex relations between the intents and the slots. 
To improve the transferability of a cross-lingual SLU system, it is crucial to employ multiple relations to achieve explicit alignment at different levels, which is ignored by the previous methods.

In this paper, we propose a novel multi-level contrastive learning (CL) framework to perform explicit alignment.
First, at the utterance level, we develop a CL scheme to enhance the intent consistency of code-switched utterances (Figure~\ref{fig:ucl}). Let $\bm{x}_i$ be an utterance in a batch of source language training data. The corresponding code-switched utterance $\bm{\hat{x}}_i$ is its positive example as although $\bm{\hat{x}}_i$ is expressed in mixed languages, it has a similar meaning to $\bm{x}_i$. Other instances ($\bm{x}_j$ and $\bm{\hat{x}}_j$, $j\neq i$), meanwhile, serve as the in-batch negative examples of $\bm{x}_i$.

Second, at the slot level, we formulate the relation between the slot values and the slots by aggregating information from multiple utterances (Figure~\ref{fig:scl}). Given each slot value in $\bm{x}_i$, the corresponding code-switched value in $\bm{\hat{x}}_i$ is selected as the positive example. We design a slot-guided value similarity for CL, which leverages the probability distributions on the slot set of the slot values to achieve semantic alignment instead of computing similarity between them directly. Furthermore, we introduce a novel algorithm to generate hard negative examples from similar slot values.

Last, at the word level, we enhance the relation between the words and their slot labels using the context in an utterance (Figure~\ref{fig:wcl}). Each word in the slot is a positive example of its slot label. We sample the words locally within the utterance as negative examples, which can be either labeled as other slot labels (type 1 negative) or out of any slots (type 2 negative). Applying CL on such positive/negative examples can strengthen the correlation between words and slot labels (through type 1 negatives) and help the model better learn the slot boundary (through type 2 negatives). 

Moreover, we propose a label-aware joint model concatenating the slot set with the utterance as the input. This is motivated by the observation that, although the languages of utterances in cross-lingual SLU are diverse, the slot set is language-invariant. By listing the slot set as the context for the utterances in different languages, the words and the slot set can attend to each other's representations in the model. The slots are \emph{implicit anchors} aligning the semantically similar words in different languages. 

We conduct extensive experiments on two benchmark datasets. The experimental results show that the proposed label-aware joint model with a multi-level CL framework significantly outperforms strong baselines. Further analysis demonstrates the effectiveness of our method.

\begin{figure*}[t]
    \centering
        \subfigure[Label-aware Joint Model]{
            \label{fig:laj}
            \includegraphics[width=0.37\textwidth]{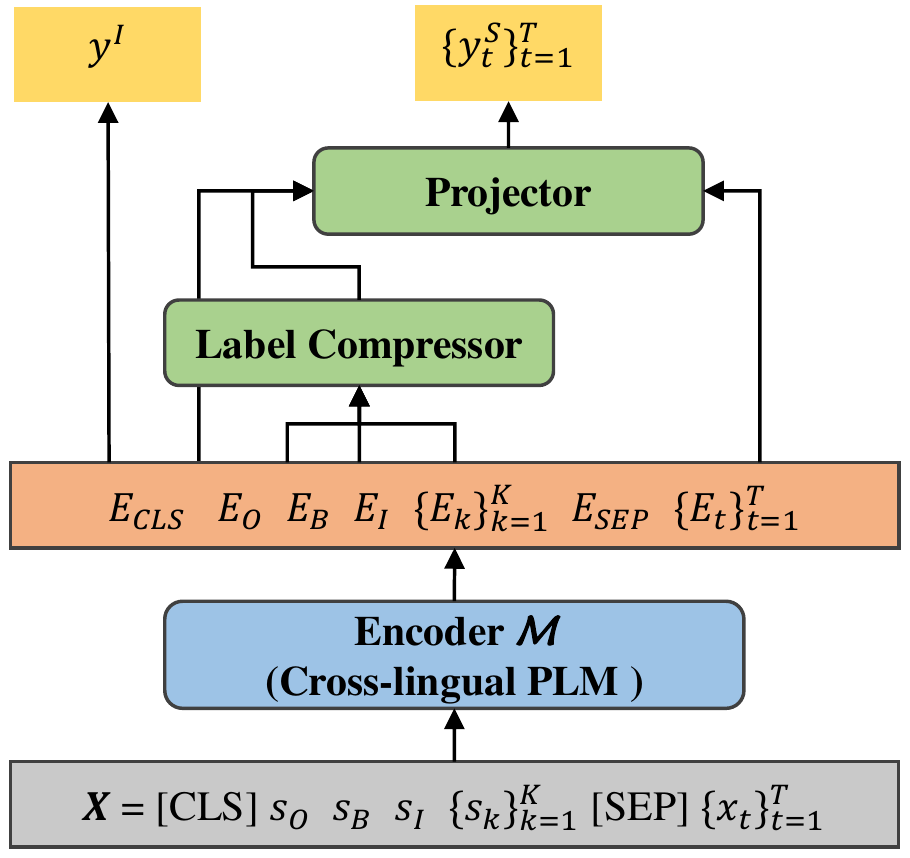}}
        \subfigure[Multi-level Contrastive Learning]{
            \label{fig:mcl}
            \includegraphics[width=0.6\textwidth]{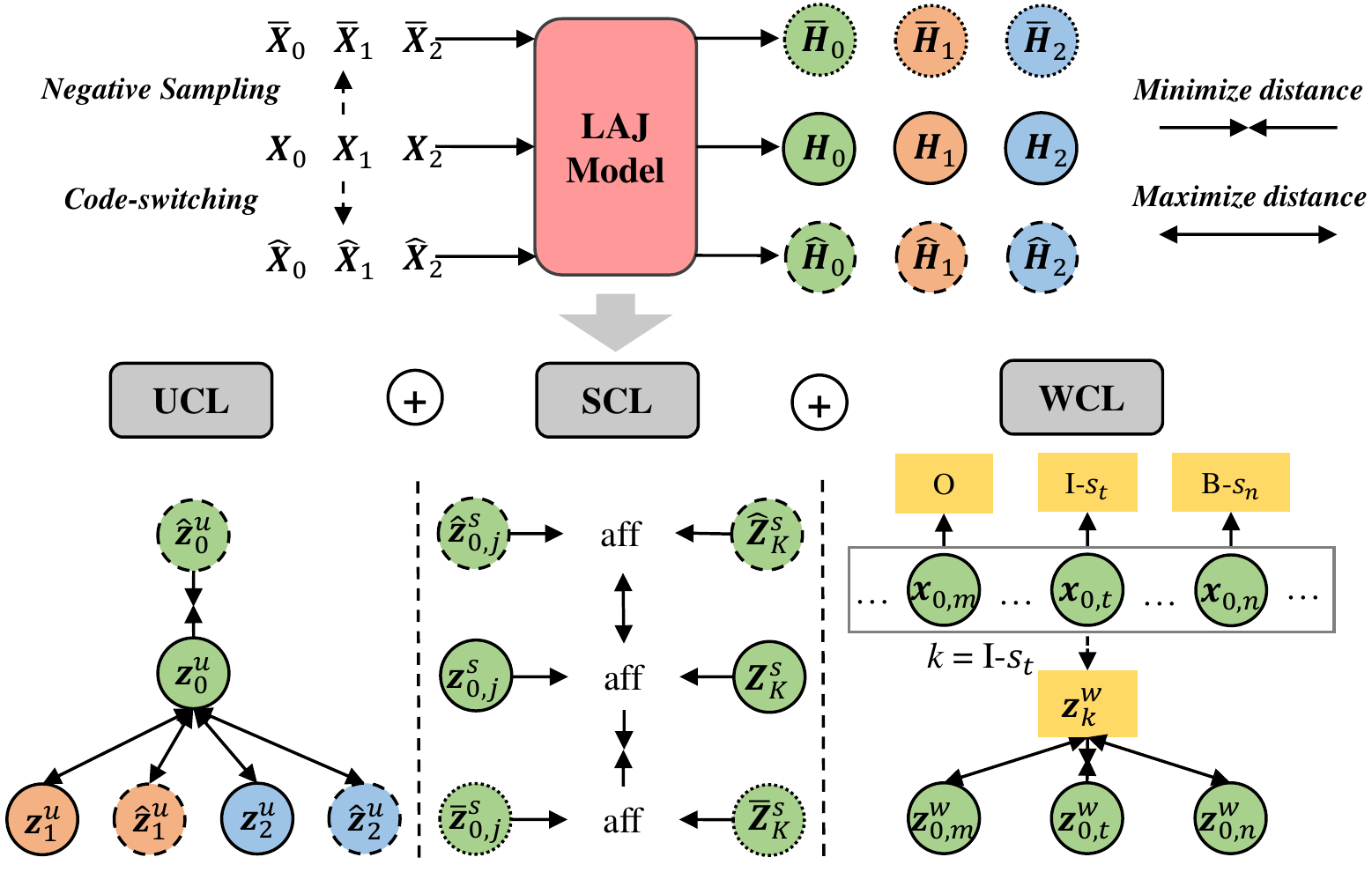}} \\
    \caption{The architecture of LAJ-MCL for cross-lingual SLU. In (b), we take the input $\bm{X}_0$ for illustration. The $\bm{x}_{0,j}$ and $\bm{x}_{0,t}$ denotes the slot value and word in the utterance $\bm{x}_0$ respectively.}
    \label{fig:laj-mcl}
\end{figure*}

\section{Related Work}
\subsection{Cross-lingual SLU}
In general, most cross-lingual SLU methods fall into two categories: model-based transfer methods and data-based transfer methods. 

The model-based transfer methods are based on cross-lingual word embeddings and contextual embeddings, such as are MUSE~\cite{lample2018word}, CoVE~\cite{mccann2017learned} and cross-lingual pre-trained language models (PLMs) including mBERT~\cite{devlin2019bert}, XLM-R~\cite{conneau2020unsupervised} and etc. A typical model-based transfer model is first fine-tuned on the source language data and then directly applied to target languages~\cite{upadhyay2018almost,schuster2019cross,li2021mtop}. More recently, additional components and training strategies have been developed.~\citet{liu2020cross} perform regularization by label sequence and adversarial training on the latent variable model~\cite{liu2019zero}.~\citet{van2021masked} propose three non-English auxiliary tasks to improve cross-lingual transfer.

The data-based transfer methods focus on building training data in target languages. Machine translation is widely adopted to translate utterances from source languages to target languages and has been shown to effectively improve model performance~\cite{schuster2019cross}.~\citet{xu2020end} propose a joint attention module for aligning the translated utterance to the slot labels to avoid label projection errors. As translated data has inherent errors and may be unavailable in low-resource languages,~\citet{liu2020attention} and~\citet{DBLP:conf/ijcai/QinN0C20} construct code-switching training data with bilingual dictionaries and fine-tune cross-lingual PLMs for implicit alignment.

\subsection{Contrastive Learning}
Contrastive learning~\cite{saunshi2019theoretical} targets at learning example representations by minimizing the distance between the positive pairs in the vector space and maximizing the distance between the negative pairs. 
In NLP, CL is first applied to learn sentence embeddings~\cite{giorgi2021declutr,gao2021simcse}. Recent studies have extended to cross-lingual PLMs.~\citet{chi2021infoxlm} unify the cross-lingual pre-training objectives by maximizing mutual information and propose a sequence-level contrastive pre-training task.~\citet{wei2021on} and~\citet{li2021multi} design a hierarchical CL by randomly sampling negative examples. Unlike the above methods using unlabeled parallel data for post-pretraining alignment,~\citet{gritta2021xeroalign} and~\citet{gritta2022crossaligner} utilize the \texttt{[CLS]} representation for utterance-level CL based on the translated data for task-specific alignment.

There is a contemporaneous work~\cite{qin2022gl} that also proposes a multi-level CL method for explicit alignment. Similarly, we both take other utterances (source and code-switched) as negative examples at the utterance level. The differences are: (1)~\citet{qin2022gl} align the words while we focus on the slot label-word relation at the word level; (2)~\citet{qin2022gl} develop the alignment between the \texttt{[CLS]} and the words at the semantic level while we propose to generate negative slot values for the slot-level CL. Further comparison is in Appendix~\ref{app:related}.

\section{Methodology}
Figure~\ref{fig:laj-mcl} illustrates our method. On the left part, the proposed Label-aware Joint Model (LAJoint) transfers the semantics of the language-invariant slot set across languages. On the right part, we develop a systematic approach to Multi-level Contrastive Learning (MCL) by novel code-switching schemes. The full framework is called LAJ-MCL.

\subsection{Label-aware Joint Model} \label{sec:joint_model}
Given an utterance $\bm{x} = \{x_t\}_{t=1}^T$ with $T$ words, the corresponding intent label and the slot label sequence are $y^I$ and $\bm{y}^S = \{y^S_t\}_{t=1}^T$, respectively.
The architecture of our label-aware joint model is shown in Figure~\ref{fig:laj}. We adopt a cross-lingual PLM as the encoder $\mathcal{M}$. The input sequence consists of three parts: (1) the three special symbols ($s_O$, $s_B$, $s_I$), i.e., the abstract labels representing outside-of-slot, beginning-of-slot, and inside-of-slot, respectively; (2) the slot set $\mathcal{S} = \{s_k\}_{k=1}^K$ corresponding to the $K$ slots in the SLU task; and (3) the utterance $\{x_t\}_{t=1}^T$. We concatenate the above three parts and add the special tokens \texttt{[CLS]} and \texttt{[SEP]}. The whole sequence is,
\begin{equation} \small
    \bm{X} = \left\{\texttt{[CLS]} s_O, s_B, s_I, \{s_k\}_{k=1}^K \texttt{[SEP]} \{x_t\}_{t=1}^T \right\}
\end{equation}

\paragraph{Embeddings for Slot Labels}
We notice that the text descriptions of slot labels often convey specific meanings. For example, the slot \texttt{fromloc} in Figure~\ref{fig:wcl} indicates that it's related to location names. Therefore, to initialize the embeddings of the abstract labels ($s_O$, $s_B$, $s_I$) and the slot set $\mathcal{S}$, the slot labels are encoded by leveraging the semantics of their text descriptions through the encoder. 
We first feed the tokens of each slot label to $\mathcal{M}$ and take the \emph{mean-pooling} over the hidden states of the bottom 3 layers to obtain the token embedding. Then the normalized \emph{mean-pooling} over each token within the slot label is utilized as the initial embedding.

Next, $\bm{X}$ is encoded by $\mathcal{M}$ to obtain the contextual embeddings for the input utterance and slot labels, i.e., $\bm{E} = \mathcal{M}(\bm{X})$, where $\bm{E} \in \mathbb{R}^{(5+K+T)\times d}$ is the representation matrix and $d$ is the dimension of the hidden states of $\mathcal{M}$.

\paragraph{Label Compressor and Projector}
There are two corresponding labels for each slot $s_k \in \mathcal{S}$. $\text{B-}s_k$ marks the beginning of the slot and $\text{I-}s_k$ indicates that a word is inside the slot. 
To learn the representation of slot labels in the BIO format~\cite{ramshaw1999text}, we use a label compressor to combine the abstract labels with the slots. First, the encoder result $\bm{E}_k$ for $s_k$ is concatenated with the result for abstract labels, i.e., $\bm{E}_B$ for $s_B$ and $\bm{E}_I$ for $s_I$, then fed to the label compressor:
\begin{align}\label{eq:compressor} \small
    \bm{E}^B_{k} &= (\bm{E}_B \| \bm{E}_k)\bm{W}_{cb} + \bm{b}_{cb} \\
    \bm{E}^I_{k} &= (\bm{E}_I \| \bm{E}_k)\bm{W}_{ci} + \bm{b}_{ci}
\end{align}
where $\bm{W}_{cb}\in \mathbb{R}^{2d\times d}$ and $\bm{W}_{ci} \in \mathbb{R}^{2d\times d}$ are weight matrices, and $\bm{b}_{cb}$ and $\bm{b}_{ci}$ are bias vectors. 

Before calculating the association between the words and the slot labels, we further apply a projector similar to~\citet{hou2020few} as below: 
\begin{align} \small
    \bm{H}_{t} &= \bm{E}_t\bm{W}_{p} + \bm{b}_{p} \\
    \bm{H}_{k} &= \bm{E}_k\bm{W}_{p} + \bm{b}_{p}
\end{align}
where $\bm{W}_p \in \mathbb{R}^{d \times d}$ is the weight matrix and $\bm{b}_{p}$ is the bias vector. 
Here, $\bm{E}_k \in \{\bm{E}_O, \{\bm{E_k}^B\}, \{\bm{E_k}^I\}\}$, where $\bm{E}_O$ is the encoder output for \emph{$s_O$}, $\{\bm{E_k}^B\}$ and $\{\bm{E_k}^I\}$ are the label compressor outputs for $B$ labels and $I$ labels. $\bm{E}_t$ is the encoder output for word $x_t$. We hope the projector learns a better representation that the semantically related words and labels can be mapped close to each other.  

\paragraph{Optimization}
For intent detection, we leverage the hidden state of $\bm{E}_{\texttt{CLS}}$ and take $\mathcal{L}_{I} = -y^I \log\bm{p}_I$ as the loss function. Here $\bm{p}_I = \mathrm{softmax}(\bm{E}_{\texttt{CLS}}\bm{W}_I + \bm{b}_I)$ is the intent classifier output, where $\bm{W}_I$ and $\bm{b}_I$ are weight matrix and bias vector, respectively.
For slot filling, the similarity between words and slot labels, $\bm{p}^S_t = \mathrm{softmax}(\bm{H}_t\{\lVert\bm{H}_k\rVert_2\}^{\mathrm{T}})$, is utilized for prediction.The loss function is formulated as $\mathcal{L}_{S} = \sum_{t=1}^{T} -y^S_t \log\bm{p}^S_t$, where $y^S_t$ is the slot label for word $x_t$. 
Last, the intent detection and slot filling are jointly optimized as:
\begin{equation}
    \mathcal{L}_J = \mathcal{L}_{I} + \mathcal{L}_{S}
\end{equation}

\subsection{Multi-level Contrastive Learning}
Here, we propose a novel framework that employs the semantic structure for explicit alignment between the source language and target languages. As shown in Figure~\ref{fig:mcl}, we apply CL at utterance, slot, and word levels to capture complex relations, including intent-utterance, slot-value, and slot label-word.

Denote by $\mathcal{D} =\{\bm{x}_i\}_{i=1}^N$ a batch of the source language training data and by $\mathcal{\hat{D}} =\{\bm{\hat{x}}_i\}_{i=1}^N$ the code-switched data, where $N$ is the batch size.

\paragraph{Utterance-level CL}
For each source utterance $\bm{x}_i$ in $\mathcal{D}$, the corresponding code-switched instance $\bm{\hat{x}}_i$ is its positive example. As shown in Figure~\ref{fig:ucl}, all the other source utterances and code-switched instances serve as the in-batch negative examples. We denote the negative example of $\bm{x}_i$ as $\bm{\bar{x}}_i$.

First, following prior studies~\cite{wei2021on,qin2022gl}, we take the encoder output of \texttt{[CLS]} as the utterance representation for $\bm{x}_i$, $\bm{\hat{x}}_i$ and $\bm{\bar{x}}_i$.
Then, we map the representations to the contrastive space by the utterance-level projection head $g_{u}(\cdot)$: 
\begin{equation}
    \bm{z}^u = g_u(\cdot) = \sigma((\cdot)\bm{W}^u_1)\bm{W}^u_2
\end{equation}
where $\cdot$ represents $\bm{e}_i$, $\bm{\hat{e}}_i$, and $\bm{\bar{e}}_i$, $\sigma$ is the ReLU activation function. The purpose is to learn better representations for the following contrastive optimization and maintain more information in $\bm{e}$. Last, the triplet loss~\cite{wang2014learning} is adopted as the utterance-level contrastive loss:
\begin{equation}
    \mathcal{L}_u(\bm{x}_i) = \max(0, f_u(\bm{z}^u_i, \bm{\hat{z}}^u_i)-f_u(\bm{z}^u_i, \bm{\bar{z}}^u_i)+r_u)
\end{equation}
where the metric function $f_u$ is $L_2$ distance and $r_u$ is the loss margin.

\paragraph{Slot-level CL}
To conduct slot-level CL, an intuitive idea is to replace each slot value with the values that frequently appear in similar slots. In this way, the model learns to map the multilingual slot values to the corresponding slots in the vector space and differentiate values for different slots. We need to address the following questions. \textbf{Q1:} Given a slot $s_k$, how to define the \emph{similar} slots and generate the negative examples? \textbf{Q2:} How to evaluate the \emph{distance} between a slot value with both its code-switched positive and negative instance?

To answer \textbf{Q1}, we describe each slot $s_k$ as its text description and the high-frequency slot values. For example, assuming that the slot $s_k$ is \texttt{alarm\_name}, it is tokenized into a list $\mathcal{A}_k$. Through the training data, we can identify some events frequently marked as \texttt{alarm\_name}, among which the top-$p_v$ frequent ones constitute a list $\mathcal{B}_k$. $\mathcal{A}_k$ and $\mathcal{B}_k$ are concatenated and fed to a sentence embedding model to get the embedding $\bm{e}_k$ for $s_k$. The similarity between the slots is then calculated by the cosine similarity between their representations. For each slot $s_k$, we define the set of hard negative values $\mathcal{V}_k$ as the union of $\mathcal{B}_{k'}$, where $\bm{s}_{k'}$ denotes the top-$p_s$ similar slots with $s_k$.

Denote by $\bm{x}_{i,j}$ the $j$-th slot value in $\bm{x}_i$ and by $\bm{\hat{x}}_{i,j}$ the corresponding code-switched positive example in $\bm{\hat{x}}_i$. 
As shown in Figure~\ref{fig:scl}, negative examples are derived by replacing each $\bm{x}_{i,j}$ by $\bm{\bar{x}}_{i,j}$ generated as follows. To maintain the consistency with the context of the positive example, the generation is conducted on $\bm{\hat{x}}_i$. Suppose the slot of $\bm{\hat{x}}_{i,j}$ is $s_k$, we sample a value from $\mathcal{V}_k$ and perform code-switching to get the hard negative example $\bm{\bar{x}}_{i,j}$. A negative utterance $\bm{\bar{x}}_i$ is then derived after replacing the values one by one. The generation algorithm is in Appendix~\ref{app:alg}.

To answer \textbf{Q2}, a basic method is calculating the cosine similarity between the slot values. However, there exists hard $\bm{\bar{x}}_{i,j}$ that are close to $\bm{x}_{i,j}$ and $\bm{\hat{x}}_{i,j}$ in the vector space but belong to different slots. Therefore, we introduce a slot-guided value similarity to focus on the slot-level semantics.

First, we apply \emph{mean-pooling} to the encoder outputs of the slot value to obtain the slot representations $\bm{e}_{i,j}$, $\bm{\hat{e}}_{i,j}$, and $\bm{\bar{e}}_{i,j}$.
Second, we evaluate the affinity of each slot value with respect to each slot and calculate the value similarity by the KL-divergence of their affinity distributions. 
To be specific, given the representations of all slots $\bm{E}^K = \{\bm{E}_k\}_{k=1}^K$ from $\mathcal{M}$, let $\bm{E}^K$, $\bm{e}_{i,j}$, $\bm{\hat{e}}_{i,j}$ and $\bm{\bar{e}}_{i,j}$ go through the slot-level projection head $g_s(\cdot)$:
\begin{equation}
    \bm{z}^s = g_s(\cdot) = \sigma((\cdot)\bm{W}^s_1)\bm{W}^s_2
\end{equation}
where $\bm{Z}^s_K = g_s(\bm{E}^K)$. The affinity is defined as $\mbox{aff}(\bm{z}^s_{i,j})$ $=\bm{z}^s_{ij}(\bm{Z}^s_K)^{\mathrm{T}}$. Similarly, we can obtain $\mbox{aff}(\bm{\hat{z}}^s_{i,j})$ and $\mbox{aff}(\bm{\bar{z}}^s_{i,j})$. Then the slot-guided slot value similarity is formulated as:
\begin{equation}
    f_s(\bm{z}^s_{i,j}, \bm{\hat{z}}^s_{i,j}) = \mathrm{KL}(\mbox{aff}(\bm{z}^s_{i,j}), \mbox{aff}(\bm{\hat{z}}^s_{i,j}))
\end{equation}
This procedure guides the model to take the probability distribution on the slot set as the semantic information of the slot value and align this knowledge between the source language and the target languages.
Finally, the slot-level contrastive loss with the margin $r_s$ is:
\begin{equation}
    \mathcal{L}_s(\bm{x}_{i,j}) =  \max(0, f_s(\bm{z}^s_{i,j}, \bm{\hat{z}}^s_{i,j})-f_s(\bm{z}^s_{i,j}, \bm{\bar{z}}^s_{i,j})+r_s)
\end{equation}

\paragraph{Word-level CL} 
Unlike the slot-level method, which aggregates information from multiple utterances, the word-level method concentrates on the context within an utterance.
Given an input $\bm{x}_i$, denote by $\bm{x}_{i,t}$ the $t$-th word with label $\bm{y}^S_{i,t}$. We consider each slot word as a positive example of its slot label. The negative examples can be sampled from the neighborhood of $\bm{x}_{i,t}$ in the utterance as shown in Figure~\ref{fig:wcl}. Suppose the negative word belongs to another slot label (type 1 negative). In this case, CL encourages the model to differentiate different slot labels based on slot type (different slot) or label transition (same slot). Furthermore, if the negative word does not belong to any slot, i.e., marked as $O$ (type 2 negative), CL improves the model sensitivity to the slot value boundary. 

To derive the negative examples $\bm{\bar{x}}_{i,t}$, the words with the same slot label as $\bm{x}_{i,t}$ are masked. For each remaining word $\bm{x}_{i,r}$, the negative sampling probability $p_r$ is based on its relative distance to $\bm{x}_{i,t}$:
\begin{align}
    p_r = \frac{q_r}{\sum_{r^{\prime}} q_{r^{\prime}}}~\mathrm{where}~q_r = \sin({\frac{1}{|r - t|}})
\end{align}

We reuse the encoding of the slot labels and words from the projector in LAJoint model. Suppose $\bm{y}^S_{i,t}=k$. The representations for $\bm{y}^S_{i,t}$, $\bm{x}_{i,t}$ and $\bm{\bar{x}}_{i,t}$, i.e., $\bm{H}_k$, $\bm{H}_{i,t}$ and $\bm{\bar{H}}_{i,t}$, go through the word-level projection head $g_w$ similar to $g_u$ and $g_s$, then obtain $\bm{z}_{k}^w$, $\bm{z}^w_{i,t}$ and $\bm{\bar{z}}^w_{i,t}$. The word-level contrastive loss is:
\begin{equation}
    \mathcal{L}_w(\bm{x}_{i,t}) =  \max(0, f_w(\bm{z}^w_{k}, \bm{z}^w_{i,t})-f_w(\bm{z}^w_{k}, \bm{\bar{z}}^w_{i,t})+r_w)
\end{equation}
where $f_w$ is cosine similarity and $r_w$ is the loss margin. In addition, our word-level method is carried out on both source and code-switched utterances.

Finally, we derive the overall training loss of LAJ-MCL as below:
\begin{equation}
    \mathcal{L} = \mathcal{L}_J + \lambda_1\mathcal{L}_u + \lambda_2\mathcal{L}_s + \lambda_3\mathcal{L}_w
\end{equation}
where $\lambda$'s are the hyper-parameters.

\begin{table*}[tb]
    \small
    \centering
    \begin{tabular}{ll|cccccc}
    \toprule
    {\multirow{2}{*}{\textbf{Data}}} & {\multirow{2}{*}{\textbf{Methods}}} & \multicolumn{3}{c}{\textbf{mBERT}} & \multicolumn{3}{c}{\textbf{XLM-R\textsubscript{base}}} \\
    &                               & Intent Acc & Slot F1 & Sem EM & Intent Acc & Slot F1 & Sem EM\\
    \midrule
    {\multirow{3}{*}{EN}} & ZSJoint\textsuperscript{*} & 87.00     & 68.08      & 38.02     & 90.94 & 66.79 & 38.85  \\
    & Ensemble-Net                                     & 87.20     & 55.78      &   -       & - & - & - \\
    & LAJoint (ours)                                   & 88.96 &  69.96 & 40.85 & 88.42 & 67.65 & 37.35 \\
    \midrule
    {\multirow{4}{*}{EN+CS}} & CoSDA-ML\textsuperscript{*} & 90.87      &  68.08     & 43.15     & 93.04 & 70.01  & 43.72  \\
    & GL-CL{\scriptsize E}F                                              & 91.95      &  \bf 80.00 & \bf 54.09 & \bf 94.05  & 74.81  & 46.35 \\
    & LAJ-MCL (ours)                                       & \bf 92.41  &  78.23     & 52.50     & 93.49  & \bf 75.69  & \bf 47.58  \\
    & \hspace{0.1cm} \emph{w/o MCL}                        & 92.01      &  76.11     & 50.37     & 91.86  & 75.33  & 46.41  \\
    \bottomrule
    \end{tabular}
    \caption{Average results of all the languages on MultiATIS++. Results with * are from our re-implementation. The full language breakdowns are shown in Appendix~\ref{app:major}.}
    \label{tab:main_atis}
\end{table*}

\begin{table}[]
    \resizebox{1.0\columnwidth}{!}{
    \begin{tabular}{ll|ccc}
    \toprule
    \textbf{Data} & \textbf{Methods} & \textbf{Intent Acc} & \textbf{Slot F1} & \textbf{Sem EM} \\
    \midrule
    {\multirow{2}{*}{EN}}    & ZSJoint\textsuperscript{*}    & 85.56 & 67.03 & 50.35  \\
                             & LAJoint (ours)                & 82.61     & 64.22     & 46.55 \\
    \midrule
    {\multirow{3}{*}{EN+CS}} & CoSDA-ML\textsuperscript{*}   & 90.72     & 73.34     & 58.77 \\
                             & LAJ-MCL (ours)                & \bf 91.04 & \bf 74.50 & \bf 60.11 \\
                             & \hspace{0.1cm} \emph{w/o MCL} & 90.67     & 73.61     & 58.92 \\
    \bottomrule
    \end{tabular}}
    \caption{Average results of all the languages on MTOP. Results with * are from our re-implementation. The full language breakdowns are shown in Appendix~\ref{app:major}.} 
    \label{tab:main_mtop}
\end{table}

\section{Experiments}
\subsection{Experiment Settings}
\paragraph{Datasets and Metrics}
We conduct our experiments on two cross-lingual SLU benchmark datasets: MultiATIS++~\cite{xu2020end} and MTOP~\cite{li2021mtop}.
\textbf{MultiATIS++} has 18 intents and 84 slots for each language and \textbf{MTOP} has in total 117 intents and 78 slots. The details of the datasets are provided in Appendix~\ref{app:datasets}.

We adopt the evaluation metrics in the previous works~\cite{xu2020end,li2021mtop} including intent detection accuracy, slot filling F1 score, and semantic exact match accuracy.

\paragraph{Implementation}
We build LAJ-MCL with the mBERT and XLM-R\textsubscript{base} from~\citet{wolf2020transformers} as the encoder. Bilingual dictionaries of MUSE are adopted for code-switching the same as~\citet{DBLP:conf/ijcai/QinN0C20}. Following the zero-shot setting, we use en training set and code-switching set for model training and en validation set for checkpoint saving. More details are described in Appendix~\ref{app:implement}.

\subsection{Baselines}
We compare our model to the following baselines.

\paragraph{ZSJoint.} We re-implement the zero-shot joint model~\cite{chen2019bert} (denoted as ZSJoint), which is trained on the en training set and directly applied to the test sets of target languages.

\paragraph{Ensemble-Net.}~\citet{razumovskaia2021crossing} propose an ensemble-style network whose predictions are the majority voting results of 8 trained single-source language models, which is a zero-shot model.

\paragraph{CoSDA-ML.}~\citet{DBLP:conf/ijcai/QinN0C20} propose a dynamic code-switching method that randomly performs multilingual token-level replacement. For a fair comparison, we use both the en training data and the code-switching data for fine-tuning.

\paragraph{GL-CL{\small E}F.}~\citet{qin2022gl} propose a global-local contrastive learning framework for explicit alignment. It is a concurrent work of this paper.

\subsection{Major Results}
Table~\ref{tab:main_atis} shows the results on MultiATIS++. 
First, CoSDA-ML and our LAJ-MCL significantly outperform ZSJoint and Ensemble-Net as code-switching (CS) helps to align the representations across languages implicitly.
Although both CoSDA-ML and LAJ-MCL apply code-switching, our framework considers the semantic structure of SLU and develops novel code-switching schemes in the multi-level CL. LAJ-MCL shows 21.7\% and 8.8\% improvements over CoSDA-ML on average semantic EM accuracy when using mBERT and XLM-R\textsubscript{base} respectively, which verifies the effectiveness of leveraging multi-level CL for explicit representation alignment. 
Second, LAJoint performs better than ZSJoint and achieves greater gains with code-switching. Based on mBERT, LAJoint beats ZSJoint with 7.4\% on EM accuracy, and it creases to 16.7\% comparing \emph{w/o MCL} with CoSDA-ML. It can be attributed that LAJoint introduces contextual label semantics. The replaced target language words are not only aligned with the source language words but also attend to the slot labels, which is a language adaptation process of the representations of slot labels. 

In Table~\ref{tab:main_mtop}, we investigate the generalization of LAJ-MCL on MTOP with XLM-R\textsubscript{base}. We find our methods can still improve the overall accuracy by 2.3\%. The results demonstrate that our framework can scale out to multiple datasets and more languages.

For CL baselines, LAJ-MCL achieves similar results to GL-CL{\small E}F and achieve the SOTA performance based on code-switching on both datasets respectively. 
We leave extending our framework to translated data for future work.

\begin{table}[tb]
    \centering
    \small
    \begin{tabular}{l|ccc}
    \toprule
    \bf Methods & \bf Intent Acc & \bf Slot F1 & \bf Sem EM \\
    \midrule
    LAJoint             & 88.96  & 69.96  & 40.85  \\
    \midrule
    \emph{- Compressor} & 89.21  & 69.64  & 40.01  \\
    \emph{- Projector}  & 88.54  & 69.69  & 40.35  \\
    \emph{- Comp\&Proj} & 87.95  & 68.74  & 39.90  \\
    \midrule
    LAJoint+CS          & 92.01 & 76.11 & 50.37 \\
    \midrule
    \emph{+ UCL}        & 92.42 & 77.28 & 51.36 \\
    \emph{+ SCL}        & 92.04 & 77.29 & 51.21 \\
    \emph{+ WCL}        & 92.18 & 77.00 & 50.89 \\
    \emph{+ UCL\&SCL}   & \bf 92.57 & 77.54 & 51.83 \\
    \emph{+ UCL\&WCL}   & 92.51 & 77.62 & 51.69 \\
    \emph{+ SCL\&WCL}   & 92.55 & 77.64 & 51.84 \\
    \emph{+ MCL}        & 92.41 & \bf 78.23 & \bf 52.50 \\
    \bottomrule
    \end{tabular}
    \caption{Ablation study of difference components. \emph{UCL}, \emph{SCL}, and \emph{WCL} denote utterance-level, slot-level, and word-level CL, respectively.}
    \label{tab:ablation}
\end{table}

\subsection{Further Analysis}
In this section, unless otherwise specified, all the methods use mBERT encoder on MutliATIS++.

\paragraph{Ablation Study}
To manifest the contribution of each component in LAJ-MCL, we conduct ablation experiments, and the average results are in Table~\ref{tab:ablation}. 

Both the label compressor and the projector play crucial roles in LAJoint. Intuitively, the label compressor learns the combination of abstract labels and slots, and then the projector learns to map the words closer to their corresponding slot labels. The average EM accuracy drops by 2.3\% without additional layers (\emph{- Comp\&Proj}).
When comparing LAJoint with ZSJoint, the clear improvements demonstrate the indispensability of our LAJoint model that leverages the language-invariant slot set to align representations implicitly. 

Furthermore, we find that adding every single CL outperforms LAJoint+CS. Specifically, \emph{UCL} encodes intent semantics and makes coarse-grained alignment between English and code-switched utterances by pulling the utterance representations close. Thus \emph{UCL} achieves the highest intent accuracy among them. \emph{SCL} and \emph{WCL} leverage label semantics to perform fine-grained alignment across and within utterances by focusing on slot-value and slot label-word relations, respectively. When combined in pairs, the coupled CL outperforms the single CL, which demonstrates \emph{UCL}, \emph{SCL}, and \emph{WCL} are complementary. 
Consequently, our MCL framework can achieve consistent improvements based on various cross-lingual PLMs.

The details about the full performance on all the languages are given in Appendix~\ref{app:ablation}.

\begin{figure}[t]
    \centering
    \includegraphics[width=1.0\columnwidth]{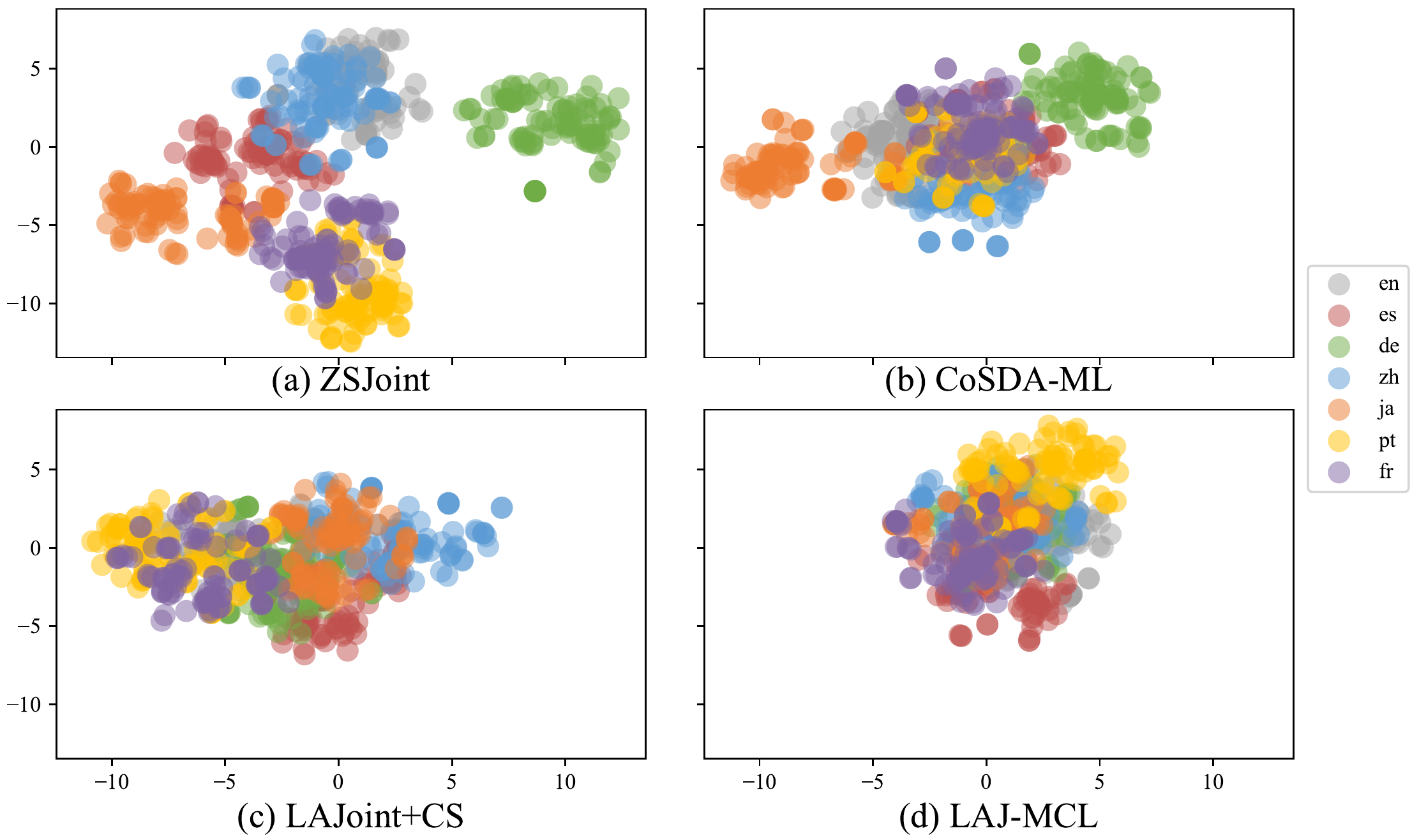}    \caption{t-SNE visualization of utterance vector space. Dots in the same color denotes the utterance representations with the same intent.}
    \label{fig:visual}
\end{figure}

\begin{table}[t]
    \centering
    \resizebox{1.0\columnwidth}{!}{
    \begin{tabular}{l|cccccccc|c}
    \toprule
    \bf Errors & es & de & zh & ja & pt & fr & hi & tr & \bf Avg. \\ \midrule
    \multicolumn{10}{l}{Method: CoSDA-ML}  \\ \midrule
    \#\emph{utterance}       & 420 & 465 & 359 & 612 & 378 & 432 & 798 & 602 & 509.5\\
    \#\emph{slot\_num}       & 231 & 342 & 201 & 171 & 228 & 202 & 308 & 386 & 258.6\\
    $^{\dagger}$\emph{slot\_type}        & 118 & 119 & 183 & 541 & 112 & 118 & 646 & 227 & 258.0\\
    $^{\dagger}$\emph{slot\_bound}    & 47  & 15  & 6   & 12  & 27  & 85  & 48  & 44 & 35.5 \\
    $^{\dagger}$\emph{slot\_both}        & 106 & 55  & 47  & 102 & 97  & 131 & 70  & 86 & 86.8 \\ \midrule
    \multicolumn{10}{l}{Method: LAJoint+CS}  \\ \midrule
    \#\emph{utterance}       & 337 & 289 & 346 & 538 & 357 & 369 & 720 & 581 & 442.1\\
    \#\emph{slot\_num}       & 159 & 179 & 186 & 147 & 215 & 208 & 296 & 331 & 215.1\\
    $^{\dagger}$\emph{slot\_type}        & 106 & 108 & 176 & 472 & 87  & 66  & 512 & 243 & 221.3 \\
    $^{\dagger}$\emph{slot\_bound}    & 84  & 14  & 12  & 8   & 41  & 85  & 93  & 60 & 49.6 \\
    $^{\dagger}$\emph{slot\_both}        & 44  & 33  & 59  & 129 & 80  & 95  & 103 & 142 & 85.6 \\ \midrule
    \multicolumn{10}{l}{Method: LAJ-MCL}  \\ \midrule
    \#\emph{utterance}       & 353 & 282 & 346 & 566 & 328 & 372 & 645 & 479 & 421.4 \\
    \#\emph{slot\_num}       & 156 & 166 & 148 & 197 & 189 & 210 & 346 & 231 & 205.4 \\
    $^{\dagger}$\emph{slot\_type}        & 106 & 110 & 227 & 459 & 189 & 70  & 349 & 247 & 219.6\\
    $^{\dagger}$\emph{slot\_bound}    & 86  & 15  & 13  & 4   & 36  & 75  & 52  & 70 & 43.9 \\
    $^{\dagger}$\emph{slot\_both}        & 77  & 45  & 34  & 85  & 101 & 110 & 78  & 59 & 73.6 \\ \bottomrule
    \end{tabular}
    }
    \caption{Error statistics of CoSDA-ML and our methods on the slot filling sub-task.}
    \label{tab:error_sts}
\end{table}

\paragraph{Visualization}
To intuitively verify whether LAJ-MCL improves the alignment of model representations between languages, we select 100 parallel utterances with the same intent from all the test sets except hi and tr of MultiATIS++. It is hard to automatically extract parallel utterances with other languages since hi and tr utterances are filtered.
Specifically, the encoder output of \texttt{[CLS]} is obtained as the utterance representation and visualized by t-SNE. The results are shown in Figure~\ref{fig:visual}.
It can be seen that there is only a small overlap between different languages in ZSJoint, i.e., the distance between different language representations is quite far. This problem is mitigated in CoSDA-ML where many dots overlaps, but there are some outliers.
For our proposed methods, LAJoint+CS has better representation alignment than CoSDA-ML. In LAJ-MCL, the overlap region is further expanded and the internal distance of each language is reduced, which fully confirms the effectiveness of explicit alignment of our MCL framework.

\begin{table}[tb]
\centering
\small
\begin{tabular}{@{}l|ccc@{}}
\toprule
Hyper-params & \multicolumn{1}{c}{$p_v$=0} & \multicolumn{1}{c}{$p_v$=10} & \multicolumn{1}{c}{$p_v$=20} \\ \midrule
$p_s$=5   & 49.78 & 50.91 & \bf 51.21 \\
$p_s$=10  & 49.75 & 51.05 & 51.15 \\ \bottomrule
\end{tabular}%
\caption{Slot-level CL hyper-parameters selection. We take LAJoint+SCL for example (Sem EM).}
\label{tab:scl_hyp_param}
\end{table}

\paragraph{Error Statistics}
We conduct error statistics on the prediction results of slot filling as shown in Table~\ref{tab:error_sts}. 
In each block, \#\emph{utterance} is the number of utterances with errors, and \#\emph{slot\_num} is the number of utterances in which the number of predicted slots are inconsistent with the ground truth. Furthermore, $^{\dagger}$\emph{slot} errors are counted from the utterances without \#\emph{slot\_num}. Specifically, $^{\dagger}$\emph{slot\_type}, $^{\dagger}$\emph{slot\_bound}, and $^{\dagger}$\emph{slot\_both} denote slot type error, slot value boundary error, and both errors, respectively. 
We have considered including the utterances with \#\emph{slot\_num} errors in the statistics. However, when the model reduces \#\emph{slot\_num}, the newly involved slot predictions affect the number of $^{\dagger}$\emph{slot} errors, leading to confusing results.

The statistics show that: 
(1) Comparing CoSDA-ML and LAJoint+CS, the statistical results are almost consistent with the experimental results. By introducing label semantics, the average numbers of \#\emph{utterance} and \#\emph{slot\_num} are significantly reduced by 13.2\% and 16.8\%, which proves that taking the slot set as the implicit anchor is effective for identifying slots in utterances on target languages.
(2) LAJ-MCL continues to reduce the number of errors. Specifically, the average numbers of $^{\dagger}$\emph{slot\_bound} and $^{\dagger}$\emph{slot\_both} drop by 11.6\% and 14.0\%, which is exactly where LAJoint+CS did not improve compared with CoSDA-ML. 
Intuitively, $^{\dagger}$\emph{slot\_type} and $^{\dagger}$\emph{slot\_bound} errors benefit from slot-level and word-level CL, respectively. Our proposed SCL semantically enhances the relationship between slots and values, and WCL improves the the model's boundary detection and slots differentiation capability. In this way, both of them also contribute to reducing \#\emph{utterance}.

\paragraph{Hyper-parameters of SCL}
To evaluate our model's sensitivity to the negative examples generation related hyper-parameters of slot-level CL, i.e., top-$p_s$ similar slots and top-$p_v$ similar values, we conduct a grid-search experiment including the ranges: $p_s=\{5,10\}$ and $p_v=\{0, 10, 20\}$ as shown in Table~\ref{tab:scl_hyp_param}. We can observe that: (1) When not concatenating $\mathcal{B}_k$ for slot representation ($p_{v}=0$), the average EM accuracy drops by 2.9\% compared with $p_{v}=20$ which confirms the effectiveness of high-frequency values; (2) Except for $p_{v}=0$, the average EM accuracy is $51.08\pm 0.017$ which indicates our slot-level CL is robust and not sensitive to the above hyper-parameters.

\begin{table*}[tb]
    \small
    \centering
    \begin{tabular}{l|cccccc}
    \toprule
    {\multirow{2}{*}{\textbf{Methods}}} & \multicolumn{3}{c}{\textbf{ATIS}} & \multicolumn{3}{c}{\textbf{SNIPS}} \\
    & Intent Acc & Slot F1 & Sem EM & Intent Acc & Slot F1 & Sem EM\\
    \midrule
    Stack-Propagation~\cite{qin2019stack}          & 97.50 & 96.10 & 88.60 & 99.00 & 97.00 & 92.90 \\
    Co-Interactive Trans~\cite{qin2021co}       & 98.00 & 96.10 & 88.80 & 97.10 & \bf 98.80 & \bf 93.10 \\
    SlotRefine~\cite{wu2020slotrefine}                 & 97.74 & 96.16 & 88.64 & 99.04 & 97.05 & 92.96 \\
    SLG~\cite{cheng2021effective}                        & 98.30 & \bf 96.20 & 88.70 & \bf 99.10 & 97.10 & \bf 93.10 \\
    LAJoint                    & \bf 98.88 & 96.08 & \bf 89.03  & 99.00  & 97.02 & 93.00 \\
    \bottomrule
    \end{tabular}
    \caption{Results on ATIS and SNIPS. All the baseline results are from the original papers.}
    \label{tab:laj_en}
\end{table*}

\begin{figure}[t]
    \centering
    \includegraphics[width=1.0\columnwidth]{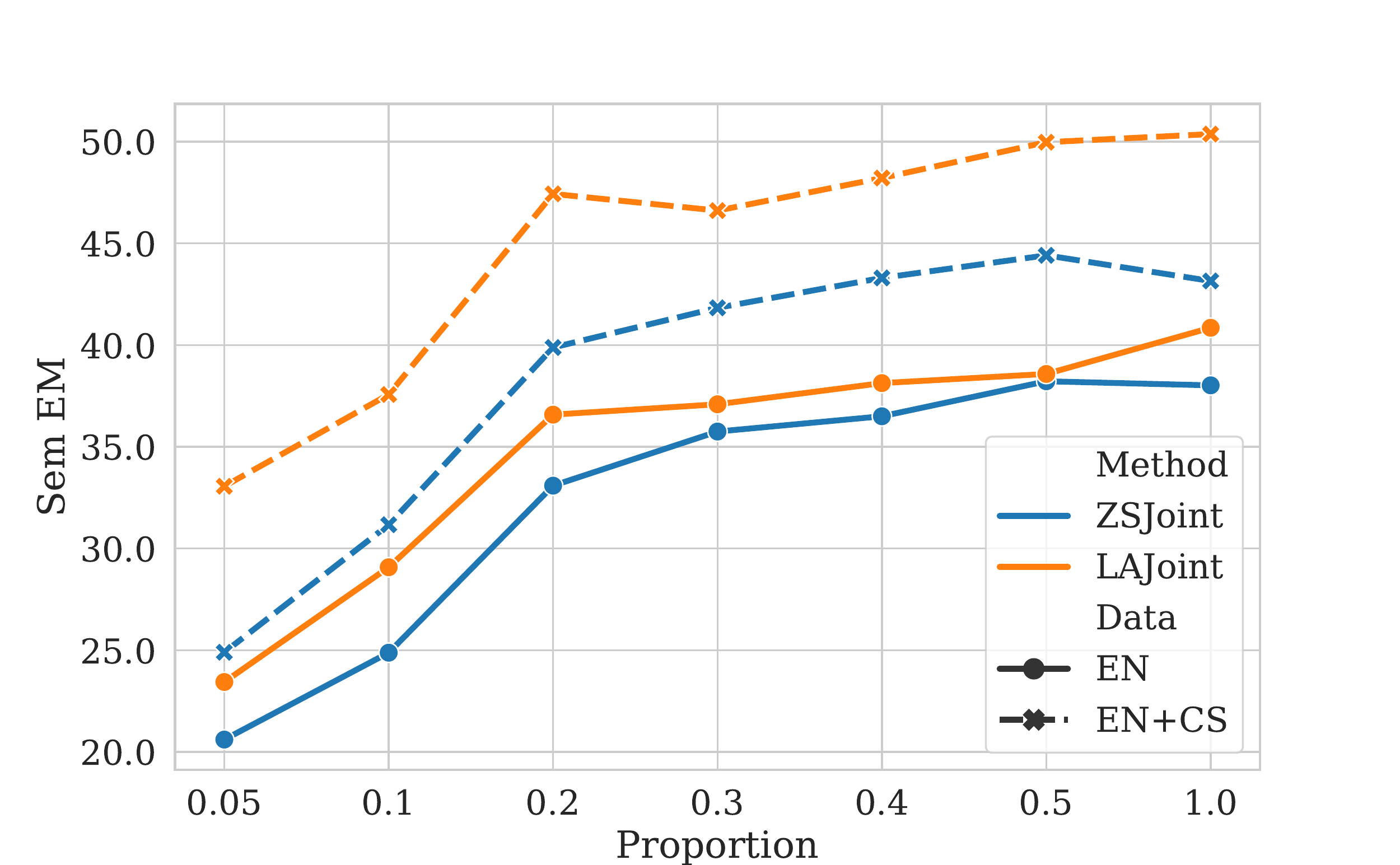}
    \caption{Comparison between ZSJoint and LAJoint by varying the proportion of training data (Sem EM).}\label{fig:prop}
\end{figure}

\paragraph{Effectiveness of LAJoint}
We conduct experiments to verify whether leveraging label semantics to facilitate the interaction between words and slots in LAJoint is effective.
In Figure~\ref{fig:prop}, LAJoint shows consistent improvements over ZSJoint with respect to different sizes and training data. The performance gap generally increases as the proportion decrease. Specifically, after applying code-switching (EN+CS), our model outperforms ZSJoint (i.e., CoSDA-ML) and increases EM accuracy by a large margin. 

We further investigate whether LAJoint works for traditional SLU tasks, including AITS~\cite{hemphill1990atis} and SNIPS~\cite{coucke2018snips}. Following the setting of current SOTA methods, we take BERT\textsubscript{base}~\cite{devlin2019bert} as the encoder. The batch size is set to 32 and 64 for ATIS and SNIPS, respectively. The learning rate is selected from \{5e-5, 6e-5, 7e-5, 8e-5, 9e-5\} and the proportion of warm-up steps is 5\%. Other details remain consistent with the multilingual experiment. As the results in Table~\ref{tab:laj_en}, LAJoint shows competitive performance compared to both autoregressive~\cite{qin2019stack,qin2021co} and non-autoregressive~\cite{wu2020slotrefine,cheng2021effective} methods. As we don't incorporate the side information such as task-interaction~\cite{qin2019stack,qin2021co} and sequential dependency~\cite{cheng2021effective}, it demonstrates that leveraging slot labels as the context of utterances is a simple and effective design.

\section{Conclusion}
In this paper, we propose a novel Label-aware Joint model (LAJoint) with a Multi-level Contrastive Learning framework (MCL) for zero-shot cross-lingual SLU. The former leverages the language-invariant slot set to transfer knowledge across languages and the latter exploits the semantic structure of SLU and develops contrastive learning based on novel code-switching schemes for explicit alignment. The results of extensive experiments demonstrate the effectiveness of our methods.

\section*{Limitations}
The main contributions of this paper are towards aligning the representations of cross-lingual PLMs implicitly and explicitly by label semantics and multi-level contrastive learning. Our methods can be extend to other cross-lingual sequence labeling tasks. Nevertheless, we summarize two limitations for further discussion and investigation of the research community:

(1) \emph{The improvement of LAJ-MCL on MTOP is not much significant as that on MultiATIS++}. MTOP has more intent labels than slot labels, and 6.5 times as many as MultiATIS++. We conjecture that it leads to a biased training process for LAJoint. In the future work, we plan to incorporate the intent labels to make full use of label semantics and achieve an unbiased training process. 

(2) \emph{The training and inference runtime of LAJ-MCL is larger than that of baselines.} The detailed results are in Table~\ref{tab:runtime}. We attribute the extra cost to the fact that LAJoint has longer input than ZSJoint, and LAJ-MCL dynamically generates negative examples in every batch. In the future work, we plan to design a new paradigm to replace the concatenation, thus reducing the requirement for GPU resources.

\section*{Acknowledgements}
Shining Liang’s research is supported by the National Natural Science Foundation of China (61976103, 61872161), the Scientific and Technological Development Program of Jilin Province (20190302029GX). Jian Pei’s research is supported in part by the NSERC Discovery Grant program. All opinions, findings, conclusions and recommendations in this paper are those of the authors and do not necessarily reflect the views of the funding agencies.

\bibliography{anthology,custom}
\bibliographystyle{acl_natbib}

\clearpage
\appendix
\section{Related Work}~\label{app:related}
Here we discuss in depth the differences from the contemporaneous work~\cite{qin2022gl}. To improve cross-lingual SLU task, although both~\citep{qin2022gl} and we propose contrastive learning frameworks for explicit alignment, the details are the same only at the sentence (utterance) level. And the differences include: (1) For the token-level CL, GL-CL{\scriptsize E}F aligns each en token with its code-switched token across sentences and take all the other tokens as negative samples. Our MCL aligns slot labels with the corresponding tokens in the en and code-switched sentences respectively. We develop a negative sampling strategy to strengthen the correlation between slot labels and tokens and help the model better learn the slot boundary; (2) GL-CL{\scriptsize E}F introduces semantic-level CL aligns the \texttt{[CLS]} token with all the other tokens in the en and code-switched sentences respectively. While the slot-level CL of MCL focuses on the annotation targets: slot values. First, negative values pool is generated for each slot. Then, we propose the slot-guided value similarity based on label semantics and align slot values across the en and code-switched sentences.

\section{Method}
\subsection{Algorithm of Slot-level CL}~\label{app:alg}
The Algorithm~\ref{alg:gen_neg} illustrates the process of generating negative examples in our slot-level CL. Here, $p_v$, $p_s$, and $N_s$ are the hyper-parameters. For $p_v$ and $p_s$, we empirically set them as 20 and 5, respectively. And for the negative examples, in order to balance the learning effectiveness and GPU memory usage, we set $N_s=2$. In our word-level CL, for the negative examples $|\{\bm{\bar{x}}_{i,t}\}| = N_w$, we usually set $N_w=2$.

\renewcommand{\algorithmicrequire}{\textbf{Input:}} 
\renewcommand{\algorithmicensure}{\textbf{Output:}}
\begin{algorithm}[t]
	\caption{: Generating Slot-level Negative Examples.} 
	\label{alg:gen_neg} 
	\begin{algorithmic}[1]
		\REQUIRE   \ 
		English utterance $\bm{x}_i$, Code-switched utterance $\bm{\hat{x}}_i$, Slot set $\mathcal{S}=\{s_k\}_{k=1}^K$.\\
		\ENSURE   \ 
		Negative utterances $|\{\bm{\bar{x}}_i\}| = N_s$.\\
		\FOR{$k=1$ to $K$}
		\STATE Tokenize $s_k$ into words and symbols to obtain $\mathcal{A}_k$
		\STATE Find the top-$p_v$ frequent slot values of $s_k$ to obtain $\mathcal{B}_k$
		\STATE Concatenating $\mathcal{A}_k$ and $\mathcal{B}_k$ as the input of MPNet provided by SentenceTransformers~\footnote{\url{https://www.sbert.net/}} to obtain the representation $\bm{e}_k$ for $s_k$
		\ENDFOR
		\FOR{$k=1$ to $K$}
		\STATE Select top-$p_s$ similar slots for $s_k$ by calculating the cosine similarity between the representations
        \STATE $\mathcal{V}_k =$ [slot values $\mathcal{B}_{k'}$ of each negative slot $s_{k'}$]
        \ENDFOR
        \FOR{each slot value $\bm{\hat{x}}_{i, j}$ in $\bm{\hat{x}}_i$}
		\STATE Suppose the slot of $\bm{\hat{x}}_{i, j}$ is $s_k$
		\STATE Randomly sample $N_s$ instances from $\mathcal{N}_k$ as negative slot values
        \STATE Replace $\bm{\hat{x}}_{i, j}$ with code-switched negative slot values iteratively to generate $\bm{\bar{x}}_{i, j}$
        \ENDFOR
	\end{algorithmic} 
\end{algorithm}

\section{Experiment Settings}
\subsection{Datasets}~\label{app:datasets}
\paragraph{MultiATIS++} is an extension of Multilingual ATIS (Table~\ref{tab:atis_sts}). Human-translated data for six languages including Spanish (es), German (de), Chinese (zh), Japanese (ja), Portuguese (pt), French (fr) are added to Multilingual ATIS which initially has Hindi (hi) and Turkish (tr). There are 4478 utterances in the train set, 500 in the valid set, and 893 in the test set, with 18 intents and 84 slots for each language. 

\paragraph{MTOP} is collected from the interactions between human and assistant systems (Table~\ref{tab:mtop_sts}). MTOP contains totally 100k+ human-translated utterances in 6 languages (English (en), German (de), Spanish (es), French (fr), Thai (th), Hindi (hi)) across 11 domains. We use the flat version divided into 70:10:20 percentage splits for train, valid and test.

\begin{table}[t] \small
    \centering
    \begin{tabular}{lccccc}
    \toprule
    \multirow{2}{*}{\textbf{Language}} & \multicolumn{3}{c}{\textbf{Utterances}} & \textbf{Intent} & \textbf{Slot} \\
    & train & valid & test & \textbf{types} & \textbf{types} \\
    \midrule
    en & 4488 & 490 & 893 & 18 & 84 \\
    es & 4488 & 490 & 893 & 18 & 84 \\
    pt & 4488 & 490 & 893 & 18 & 84 \\
    de & 4488 & 490 & 893 & 18 & 84 \\
    fr & 4488 & 490 & 893 & 18 & 84 \\
    zh & 4488 & 490 & 893 & 18 & 84 \\
    ja & 4488 & 490 & 893 & 18 & 84 \\
    hi & 1440 & 160 & 893 & 17 & 75 \\
    tr & 578 & 60 & 715 & 17 & 71 \\
    \bottomrule
    \end{tabular}
    \caption{Statistics of MultiATIS++}
    \label{tab:atis_sts}
\end{table}

\begin{table}[t] \small
    \resizebox{1.0\columnwidth}{!}{
    \begin{tabular}{cccccccc}
    \toprule
    \multicolumn{6}{c}{\textbf{Number of utterances (train/valid/test)}} & \textbf{Intent} & \textbf{Slot} \\
    en & de & fr & es & hi & th & \textbf{types} & \textbf{types} \\
    \midrule
    22288 & 18788 & 16584 & 15459 & 16131 & 15195 & 117 & 78 \\
    \bottomrule
    \end{tabular}}
    \caption{Statistics of MTOP}
    \label{tab:mtop_sts}
\end{table}

\subsection{Implement Details}~\label{app:implement}
For code-switching, the sentence replacement ratio is set to $1.0$ and the word replacement ratio is set to $0.9$. We set the batch size $N$ to 32 and train the model for 20 epochs. We apply the AdamW optimizer with the linear scheduler. 
We select the best hyper-parameters by grid search including the ranges: learning rate of the encoder \{1e-5, 2e-5, 3e-5, 4e-5, 5e-5\}; weight decay \{0, 1e-3\}; margin $r$'s in triplet loss $r$'s \{0.1, 0.3, 0.5, 0.7\}; loss coefficient $\lambda$'s \{0.3, 0.5, 0.7, 1.0\}. The learning rate of the intent classifier and contrastive learning projection heads is 1e-3. The proportion of warm-up steps is 10\%.

Following the zero-shot setting, we fine-tune the model on en training set and use en validation set for the hyper-parameters search. The best model checkpoint is decided by the semantic EM accuracy on en validation set. All the experiments are conducted on NVIDIA A100 and A6000 GPUs with NVIDIA's Automatic Mixed Precision. Our code is based on PyTorch and Transformers~\footnote{https://github.com/huggingface/transformers}.

\section{Further Discussions}
\subsection{Full Major Results}~\label{app:major}
The full comparison results on MultiATIS++ (Table~\ref{tab:main_atis_mbert},~\ref{tab:main_atis_xlmr}) and MTOP (Table~\ref{tab:main_mtop_xlmr}).

\subsection{Full Ablation Study}~\label{app:ablation}
The full ablation results on MultiATIS++ are shown in Table~\ref{tab:ablation}.
We observe that:
(1) Different level CL methods show different sensitivity to the target languages. For example, \emph{WCL} outperforms \emph{SCL} on western languages, i.e., es, de, and fr, but significantly falls behind on ja, hi, and tr. In real scenarios, one can flexibly combine them according to the target languages.
(2) The main performance improvement comes from the slot filling task. \emph{MCL} shows 0.4\% and 2.8\% average improvements over LAJoint+CS on intent detection accuracy and slot filling F1 score, respectively.
(3) Even though single CL, coupled CL, or MCL can not always perform better than LAJoint+CS on every target language, they achieve consistent improvement in average results of the three metrics, which indicates the generalization of our framework.

\subsection{Case Study}~\label{app:case}
Table~\ref{tab:case} lists several examples to illustrate the rationale behind our MCL method. In the first case, \textbf{mittag} means ``noon" in English, and \texttt{depart\_time.time} is the most frequently misclassified slot of \texttt{depart\_time.period\_of\_day} according to our empirical study on the results of the baseline methods. Such errors can be addressed by our slot-level CL, which replaces the words in a slot span with the words frequently in similar slots. 

In the second case, \textbf{más temprano} means ``earlier" in English. By word-level CL, the model reduces the error in slot boundary detection and changes from beginning-of-slot (\emph{B}) to inside-of-slot (\emph{I}). 

For the last case, \textbf{tacoma havaalani} means ``tacoma airport" in English. Our method learns to extend the slot value (through WCL). Moreover, the slot type is further corrected from \texttt{city\_name} to \texttt{airport\_name}, which can be attributed to SCL. This case demonstrates the effectiveness of applying multi-level contrastive learning jointly.

\begin{table}[tb]
\small
\centering
\begin{tabular}{l|cc}
\toprule
\textbf{Methods} & \textbf{training} & \textbf{inference} \\ \midrule
ZSJoint          & 14                & 9                  \\
CoSDA-ML         & 15                & 9                  \\
LAJoint          & 18                & 16                 \\
LAJoint+CS       & 30                & 16                 \\
LAJ-MCL          & 65                & 16                 \\ \bottomrule
\end{tabular}
\caption{Comparison of training and inference runtime (second/epoch).}
\label{tab:runtime}
\end{table}

\begin{table*}[htb]
    \resizebox{1.0\textwidth}{!}{
    \begin{tabular}{p{0.4\textwidth}|c|c|c}
    \toprule
    \bf Case & \bf w/o MCL Result & \bf Method & \bf MCL Result \\
    \midrule
    (1) Ich brauche Fluginformationen für einen Flug von Indianapolis nach Cleveland , der am Dienstag \textbf{mittag} abfliegt & \emph{\color{blue}{B-}}\texttt{{\color{blue}{depart\_time.}}{\color{red}{time}}} & \emph{+ SCL} & \emph{\color{blue}{B-}}\texttt{\color{blue}{depart\_time.period\_of\_day}} \\
    (2) cuál es el vuelo \textbf{más temprano} entre baltimore y oakland con desayuno  & \emph{\color{blue}{B-}}\texttt{\color{blue}{flight\_mod}}  \emph{\color{red}{B-}}\texttt{\color{blue}{flight\_mod}} & \emph{+ WCL} & \emph{\color{blue}{B-}}\texttt{\color{blue}{flight\_mod}}  \emph{\color{blue}{I-}}\texttt{\color{blue}{flight\_mod}}  \\
    (3) \textbf{tacoma havaalani} , havalanindan sehir merkezine ulasim sagliyor mu ? & \emph{\color{blue}{B-}}\texttt{\color{red}{city\_name}} \emph{\color{red}{O}} & \emph{+ WCL\&SCL} & \emph{\color{blue}{B-}}\texttt{\color{blue}{airport\_name}} \emph{\color{blue}{I-}}\texttt{\color{blue}{airport\_name}} \\
    \bottomrule
    \end{tabular}
    }
    \caption{Case study of our MCL on MutliATIS++. \textbf{Bold} span in the case is the target slot value. {\color{red}{Red}} and {\color{blue}{Blue}} indicate the false and true parts in the results, respectively.}
    \label{tab:case}
\end{table*}

\begin{table*}[tb]
    \centering
    \small
    \begin{tabular}{l|ccccccccc|c}
    \toprule
    \bf Intent Acc & en & es & de & zh & ja & pt & fr & hi & tr & \bf Avg. \\
    \midrule
    ZSJoint*      & 98.54 & 93.28 & 90.48 & 84.55 & 76.59 & 94.62 & 94.51 & 77.15 & 73.29 & 87.00 \\
    Ensemble-Net  & 90.26 & 96.64 & 92.50 & 84.99 & 77.04 & 95.30 & 95.18 & 77.88 & 75.04 & 87.20 \\
    LAJoint       & 98.54 & 96.30 & 93.17 & 89.25 & 83.31 & 95.41 & 95.97 & 82.53 & 66.15 & 88.96 \\
    \midrule
    CoSDA-ML*     & 97.98 & 95.07 & 95.07 & 91.04 & 85.67 & 95.18 & 95.97 & 84.88 & 76.92 & 90.87 \\
    GL-CL{\scriptsize E}F       & 98.77 & 97.05 & 97.53 & 87.68 & 82.84 & 96.08 & 97.72 & 86.00 & 83.92 & 91.95 \\
    LAJ-MCL       & 98.77 & 98.10 & 98.10 & 89.03 & 81.86 & 97.09 & 98.77 & 84.54 & 85.45 & \bf 92.41 \\
    \hspace{0.1cm} \emph{w/o MCL} & 98.32 & 97.87 & 97.31 & 91.60 & 86.67 & 96.86 & 97.76 & 88.58 & 73.15 & 92.01 \\
    \midrule
    \bf Slot F1   & en & es & de & zh & ja & pt & fr & hi & tr & \bf Avg. \\
    \midrule
    ZSJoint*      & 95.20 & 76.52 & 74.79 & 66.91 & 70.10 & 72.56 & 74.25 & 52.73 & 29.66 & 68.08 \\
    Ensemble-Net  & 85.05 & 77.56 & 82.75 & 37.29 & 9.44  & 74.00 & 76.19 & 14.14 & 45.63 & 55.78 \\
    LAJoint       & 95.80 & 80.69 & 76.63 & 67.24 & 74.47 & 72.20 & 76.23 & 54.22 & 32.12 & 69.96 \\
    \midrule
    CoSDA-ML*     & 95.32 & 80.82 & 79.63 & 80.40 & 65.69 & 79.30 & 79.21 & 49.29 & 50.53 & 73.36 \\
    GL-CL{\scriptsize E}F       & 95.39 & 85.22 & 86.30 & 77.61 & 73.12 & 81.83 & 84.31 & 70.34 & 65.85 & \bf 80.00 \\
    LAJ-MCL       & 96.02 & 83.03 & 86.59 & 82.00 & 68.52 & 81.49 & 82.11 & 61.04 & 65.20 & 78.23 \\
    \hspace{0.1cm} \emph{w/o MCL} & 95.39 & 85.85 & 86.13 & 81.35 & 69.45 & 81.17 & 82.42 & 54.01 & 49.24 & 76.11 \\
    \midrule
    \bf Semantic EM & en & es & de & zh & ja & pt & fr & hi & tr & \bf Avg. \\
    \midrule
    ZSJoint      & 87.23 & 44.46 & 41.43 & 30.80 & 33.59 & 43.90 & 43.67 & 16.01 & 1.12  & 38.02 \\
    LAJoint      & 88.24 & 43.56 & 47.03 & 38.86 & 44.46 & 40.99 & 41.77 & 20.94 & 1.82  & 40.85 \\
    \midrule
    CoSDA-ML     & 87.23 & 50.28 & 45.35 & 55.10 & 26.32 & 55.21 & 49.38 & 8.29  & 11.61 & 43.15 \\
    GL-CL{\scriptsize E}F      & 88.02 & 59.53 & 66.03 & 50.62 & 41.42 & 60.43 & 57.02 & 34.83 & 28.95 & \bf 54.09 \\
    LAJ-MCL      & 89.81 & 59.13 & 67.75 & 54.76 & 29.34 & 61.93 & 57.56 & 23.29 & 28.95 & 52.50 \\
    \hspace{0.1cm} \emph{w/o MCL} & 87.46 & 61.03 & 66.97 & 56.44 & 34.83 & 58.68 & 57.56 & 16.35 & 13.99 & 50.37 \\
    \bottomrule
    \end{tabular}
    \caption{MultiATIS++ results as average Intent Detection Accuracy/Slot Filling F1 score/Semantic Exact Match Accuracy (mBERT encoder). Results with * are from our re-implementation.}
    \label{tab:main_atis_mbert}
\end{table*}

\begin{table*}[t]
    \centering
    \small
    \begin{tabular}{l|ccccccccc|c}
    \toprule
    \bf Intent Acc & en & es & de & zh & ja & pt & fr & hi & tr & \bf Avg. \\
    \midrule
    ZSJoint*  & 98.99 & 98.10 & 92.83 & 87.91 & 84.43 & 93.95 & 97.09 & 87.12 & 78.04 & 90.94 \\
    LAJoint   & 98.77 & 92.72 & 89.81 & 86.45 & 81.97 & 93.73 & 92.61 & 84.66 & 75.10 & 88.42 \\
    \midrule
    CoSDA-ML* & 98.99 & 98.99 & 98.32 & 89.70 & 83.76 & 97.98 & 98.66 & 89.70 & 81.26 & 93.04 \\
    GL-CL{\scriptsize E}F   & 98.66 & 98.04 &98.43 &91.38	&88.83	&97.76	&97.85	&93.84	&81.68 & \bf 94.05 \\
    LAJ-MCL   & 98.77 & 98.88 & 98.32 & 90.59 & 86.23 & 97.31 & 97.98 & 91.38 & 81.96 & 93.49 \\
    \hspace{0.1cm} \emph{w/o MCL} & 98.88 & 98.88 & 97.98 & 84.99 & 85.11 & 93.62 & 98.54 & 89.14 & 79.58 & 91.86 \\
    \midrule
    \bf Slot F1 & en & es & de & zh & ja & pt & fr & hi & tr & \bf Avg. \\
    \midrule
    ZSJoint*  & 95.32 & 81.55 & 82.12 & 68.92 & 39.77 & 79.20 & 78.64 & 39.83 & 35.73 & 66.79 \\
    LAJoint   & 95.87 & 77.29 & 79.89 & 70.56 & 49.49 & 76.76 & 77.77 & 45.74 & 35.50 & 67.65 \\
    \midrule
    CoSDA-ML* & 95.32 & 84.98 & 83.92 & 77.74 & 42.40 & 80.50 & 81.13 & 41.11 & 42.44 & 70.01 \\
    GL-CL{\scriptsize E}F   & 95.88 & 82.47 &84.91	&80.5	&55.57	&77.27	&80.99	&61.11	&54.55 &74.81 \\
    LAJ-MCL   & 95.87 & 83.10 & 83.88 & 79.55 & 64.30 & 79.31 & 81.43 & 54.96 & 58.80 &\bf 75.69 \\
    \hspace{0.1cm} \emph{w/o MCL} & 95.35 & 84.87 & 81.46 & 80.78 & 67.72 & 79.10 & 81.37 & 54.66 & 52.67 & 75.33 \\
    \midrule
    \bf Semantic EM & en & es & de & zh & ja & pt & fr & hi & tr & \bf Avg. \\
    \midrule
    ZSJoint  & 88.24 & 52.18 & 57.89 & 30.01 & 4.59  & 54.20 & 52.41 & 7.05  & 3.08 & 38.85 \\
    LAJoint  & 88.91 & 43.23 & 49.94 & 31.13 & 12.43 & 49.27 & 47.93 & 11.31 & 1.96 & 37.35 \\
    \midrule
    CoSDA-ML & 88.24 & 60.47 & 62.93 & 45.24 & 6.72  & 58.01 & 57.78 & 6.27  & 7.41 & 43.72 \\
    GL-CL{\scriptsize E}F  & 88.24 & 53.51 & 64.91 &52.07	&13.77	&52.35	& 58.28	& 19.49	&14.55	&46.35 \\
    LAJ-MCL  & 88.58 & 57.22 & 55.99 & 53.75 & 27.88 & 55.10 & 55.66 & 12.09 & 21.96 & \bf 47.58 \\
    \hspace{0.1cm} \emph{w/o MCL} & 88.24 & 58.57 & 50.06 & 51.40 & 30.46 & 52.86 & 58.34 & 14.22 & 13.57 & 46.41 \\
    \bottomrule
    \end{tabular}
    \caption{MultiATIS++ results as average Intent Detection Accuracy/Slot Filling F1 score/Semantic Exact Match Accuracy (XLM-R\textsubscript{base} encoder). Results with * are from our re-implementation.}
    \label{tab:main_atis_xlmr}
\end{table*}

\begin{table*}[t]
    \centering
    \small
    \begin{tabular}{l|cccccc|c}
    \toprule
    \bf Intent Acc & en & es & fr & de & hi & th & \bf Avg. \\
    \midrule
    ZSJoint   & 96.83 & 87.86 & 83.12 & 83.91 & 79.74 & 81.92 & 85.56 \\
    LAJoint   & 96.53 & 86.66 & 83.65 & 76.61 & 75.33 & 76.85 & 82.61 \\
    \midrule
    CoSDA-ML  & 96.92 & 94.16 & 91.23 & 92.76 & 80.75 & 88.50 & 90.72 \\
    LAJ-MCL   & 96.83 & 94.20 & 92.52 & 93.46 & 82.43 & 86.80 & \bf 91.04 \\
    \hspace{0.1cm} \emph{w/o MCL} & 96.85 & 93.86 & 91.79 & 92.76 & 81.93 & 86.84 & 90.67 \\
    \midrule
    \bf Slot F1 & en & es & fr & de & hi & th & \bf Avg. \\
    \midrule
    ZSJoint   & 91.88 & 70.93 & 72.94 & 67.16 & 49.88 & 49.37 & 67.03 \\
    LAJoint   & 91.58 & 68.60 & 68.46 & 61.14 & 49.72 & 45.85 & 64.22 \\
    \midrule
    CoSDA-ML  & 91.43 & 78.22 & 78.17 & 77.68 & 57.27 & 57.24 & 73.34  \\
    LAJ-MCL   & 91.78 & 78.11 & 78.16 & 78.35 & 61.79 & 58.82 & \bf 74.50 \\
    \hspace{0.1cm} \emph{w/o MCL} & 91.30 & 78.27 & 77.05 & 77.80 & 59.45 & 57.56 & 73.61 \\
    \midrule
    \bf Semantic EM & en & es & fr & de & hi & th & \bf Avg. \\
    \midrule
    ZSJoint   & 84.65 & 54.70 & 52.02 & 46.18 & 32.20 & 32.33 & 50.35 \\
    LAJoint   & 84.34 & 49.90 & 46.45 & 39.19 & 31.52 & 27.92 & 46.55 \\
    \midrule
    CoSDA-ML  & 84.02 & 65.24 & 60.82 & 62.78 & 38.72 & 41.01 & 58.77 \\
    LAJ-MCL   & 84.59 & 65.68 & 63.98 & 63.57 & 40.59 & 42.28 & \bf 60.11 \\
    \hspace{0.1cm} \emph{w/o MCL} & 83.81 & 65.28 & 61.17 & 60.52 & 41.91 & 40.83 & 58.92 \\
    \bottomrule
    \end{tabular}
    \caption{MTOP results as average Intent Detection Accuracy/Slot Filling F1 score/Semantic Exact Match Accuracy (XLM-R\textsubscript{base} encoder). Results with * are from our re-implementation.}
    \label{tab:main_mtop_xlmr}
\end{table*}

\begin{table*}[tb]
\small
\centering
\begin{tabular}{l|ccccccccc|c}
\toprule
\textbf{Intent Acc} & en & es    & de    & zh    & ja    & pt    & fr    & hi    & tr    & \textbf{Avg.} \\ \midrule
LAJoint                    & 98.54 & 96.30 & 93.17 & 89.25 & 83.31 & 95.41 & 95.97 & 82.53 & 66.15 & 88.96         \\ \midrule
\emph{-Compressor}         & 98.43 & 96.98 & 89.59 & 88.13 & 83.87 & 94.40 & 94.96 & 81.86 & 74.69 & 89.21         \\
\emph{-Projector}          & 98.32 & 96.30 & 91.27 & 89.70 & 81.97 & 93.17 & 97.42 & 83.99 & 64.76 & 88.54         \\
\emph{-Comp\&Proj}         & 98.10 & 96.19 & 93.17 & 88.24 & 84.77 & 96.08 & 95.86 & 84.43 & 54.69 & 87.95         \\ \midrule
LAJoint+CS                 & 98.32 & 97.87 & 97.31 & 91.60 & 86.67 & 96.86 & 97.76 & 88.58 & 73.15 & 92.01         \\ \midrule
\emph{+UCL}                & 98.43 & 97.65 & 96.98 & 90.93 & 88.13 & 96.98 & 98.54 & 87.12 & 77.06 & 92.42         \\
\emph{+SCL}                & 97.98 & 97.09 & 97.42 & 91.83 & 86.90 & 96.86 & 98.10 & 88.91 & 73.29 & 92.04         \\
\emph{+WCL}                & 98.32 & 97.65 & 97.54 & 90.03 & 88.58 & 96.64 & 97.98 & 87.35 & 75.52 & 92.18         \\
\emph{+UCL\&SCL}           & 98.43 & 97.31 & 97.31 & 91.71 & 89.36 & 96.08 & 98.32 & 88.35 & 76.22 & \bf 92.57         \\
\emph{+UCL\&WCL}           & 98.43 & 97.98 & 97.54 & 90.15 & 85.67 & 96.75 & 98.10 & 86.45 & 81.54 & 92.51         \\
\emph{+SCL\&WCL}           & 98.43 & 97.98 & 97.87 & 92.27 & 85.78 & 96.86 & 98.43 & 87.01 & 78.32 & 92.55         \\
\emph{+MCL}                & 98.77 & 98.10 & 98.10 & 89.03 & 81.86 & 97.09 & 98.77 & 84.54 & 85.45 & 92.41 \\ \midrule
\textbf{Slot F1}    & en & es    & de    & zh    & ja    & pt    & fr    & hi    & tr    & \textbf{Avg.} \\ \midrule
LAJoint                    & 95.80 & 80.69 & 76.63 & 67.24 & 74.47 & 72.20 & 76.23 & 54.22 & 32.12 & 69.96         \\ \midrule
\emph{-Compressor}         & 95.59 & 77.10 & 76.36 & 70.50 & 73.83 & 70.30 & 73.42 & 50.56 & 39.07 & 69.64         \\
\emph{-Projector}          & 95.75 & 76.18 & 77.56 & 69.21 & 71.82 & 74.90 & 76.64 & 49.98 & 35.18 & 69.69         \\
\emph{-Comp\&Proj}         & 95.57 & 75.50 & 74.26 & 69.16 & 73.97 & 72.96 & 73.67 & 56.38 & 27.23 & 68.74         \\ \midrule
LAJoint+CS                 & 95.39 & 85.85 & 86.13 & 81.35 & 69.45 & 81.17 & 82.42 & 54.01 & 49.24 & 76.11         \\ \midrule
\emph{+UCL}                & 95.83 & 84.03 & 84.70 & 82.28 & 76.10 & 80.99 & 81.45 & 55.00 & 55.18 & 77.28         \\
\emph{+SCL}                & 95.61 & 84.22 & 85.11 & 81.26 & 76.20 & 81.25 & 81.79 & 56.58 & 53.63 & 77.29         \\
\emph{+WCL}                & 95.73 & 83.83 & 85.15 & 82.50 & 72.85 & 80.34 & 82.14 & 55.74 & 54.68 & 77.00         \\
\emph{+UCL\&SCL}           & 95.81 & 85.71 & 84.72 & 82.23 & 73.82 & 81.70 & 81.41 & 57.10 & 55.35 & 77.54         \\
\emph{+UCL\&WCL}           & 95.76 & 84.43 & 84.39 & 81.63 & 74.39 & 81.91 & 82.39 & 55.82 & 57.88 & 77.62         \\
\emph{+SCL\&WCL}           & 95.85 & 83.15 & 83.69 & 81.45 & 73.48 & 80.60 & 78.35 & 62.07 & 60.12 & 77.64         \\
\emph{+MCL}                & 96.02 & 83.03 & 86.59 & 82.00 & 68.52 & 81.49 & 82.11 & 61.04 & 65.20 & \bf 78.23  \\ \midrule
\textbf{Sem EM}     & en & es    & de    & zh    & ja    & pt    & fr    & hi    & tr    & \textbf{Avg.} \\ \midrule
LAJoint                    & 88.24 & 52.74 & 50.84 & 34.60 & 43.23 & 42.89 & 39.31 & 18.48 & 2.10  & 40.85         \\ \midrule
\emph{-Compressor}         & 88.24 & 44.34 & 46.25 & 36.95 & 44.68 & 40.31 & 37.96 & 15.45 & 5.87  & 40.01         \\
\emph{-Projector}          & 88.24 & 43.78 & 46.36 & 36.73 & 38.41 & 42.55 & 48.04 & 13.55 & 5.45  & 40.35         \\
\emph{-Comp\&Proj}         & 88.58 & 40.87 & 44.34 & 37.63 & 44.23 & 42.67 & 37.07 & 21.50 & 2.24  & 39.90         \\ \midrule
LAJoint+CS                 & 87.46 & 61.03 & 66.97 & 56.44 & 34.83 & 58.68 & 57.56 & 16.35 & 13.99 & 50.37         \\ \midrule
\emph{+UCL}                & 88.58 & 59.91 & 65.96 & 55.43 & 40.20 & 58.23 & 52.97 & 21.39 & 19.58 & 51.36         \\
\emph{+SCL}                & 89.03 & 55.43 & 61.59 & 58.79 & 41.88 & 57.89 & 45.69 & 27.10 & 23.50 & 51.21         \\
\emph{+WCL}                & 88.80 & 58.23 & 65.29 & 58.45 & 39.53 & 56.33 & 57.78 & 15.45 & 18.18 & 50.89         \\
\emph{+UCL\&SCL}           & 87.57 & 53.53 & 63.83 & 55.99 & 50.06 & 56.66 & 52.41 & 28.11 & 18.32 & 51.83         \\
\emph{+UCL\&WCL}           & 88.69 & 55.77 & 66.29 & 55.54 & 44.01 & 58.68 & 55.88 & 21.84 & 18.18 & 51.69         \\
\emph{+SCL\&WCL}           & 89.03 & 58.45 & 63.61 & 57.11 & 43.45 & 58.68 & 54.98 & 21.28 & 20.00 & 51.84         \\
\emph{+MCL}                & 89.81 & 59.13 & 67.75 & 54.76 & 29.34 & 61.93 & 57.56 & 23.29 & 28.95 & \bf 52.50 \\\bottomrule
\end{tabular}%
\caption{Ablation study of difference components on MutliATIS++ (mBERT encoder).}
\label{tab:ablation}
\end{table*}

\end{document}